\documentclass[10pt, conference, compsocconf]{IEEEtran}

\usepackage{graphics} % for pdf, bitmapped graphics files
\usepackage{epsfig} % for postscript graphics files
\usepackage{times} % assumes new font selection scheme installed
\usepackage{amsmath} % assumes amsmath package installed
\usepackage{amssymb}  % assumes amsmath package installed
\usepackage{breqn}
\usepackage{bm}
\usepackage{algorithm}
\usepackage{algpseudocode}
\usepackage{physics}
\usepackage[export]{adjustbox}
\usepackage{caption}

\usepackage{bibspacing}
\ifCLASSINFOpdf
\begin{document}
\title{Learning a Bias Correction for Lidar-only Motion Estimation}


\author{\IEEEauthorblockN{Tim Y. Tang, David J. Yoon}
\IEEEauthorblockA{Institute for Aerospace Studies\\
University of Toronto\\
Toronto, Canada\\
Email: \{tim.tang, david.yoon\}\\ @robotics.utias.utoronto.ca}
\and
\IEEEauthorblockN{Fran\c{c}ois Pomerleau}
\IEEEauthorblockA{Dept. Computer Science and Software Engineering\\
Laval University\\
Quebec City, Canada\\
francois.pomerleau@ift.ulaval.ca}
\and
\IEEEauthorblockN{Timothy D. Barfoot}
\IEEEauthorblockA{Institute for Aerospace Studies\\
University of Toronto\\
Toronto, Canada\\
Email: tim.barfoot@utoronto.ca}
}

\maketitle

\begin{abstract}
This paper presents a novel technique to correct for bias in a classical estimator using a learning approach. We apply a learned bias correction to a lidar-only motion estimation pipeline. Our technique trains a Gaussian process (GP) regression model using data with ground truth. The inputs to the model are high-level features derived from the geometry of the point-clouds, and the outputs are the predicted biases between poses computed by the estimator and the ground truth. The predicted biases are applied as a correction to the poses computed by the estimator. 

Our technique is evaluated on over 50\,km of lidar data, which includes the KITTI odometry benchmark and lidar datasets collected around the University of Toronto campus. After applying the learned bias correction, we obtained significant improvements to lidar odometry in all datasets tested. We achieved around 10\% reduction in errors on all datasets from an already accurate lidar odometry algorithm, at the expense of only less than 1\% increase in computational cost at run-time.

\end{abstract}  


% \fi
\IEEEpeerreviewmaketitle

\section{Introduction}
\par 
Mobile robots rely on on-board sensors such as lidar or camera for accurate motion estimation. Lidars are particularly useful as they are relatively unaffected by lighting conditions, and have a larger field-of-view than cameras. In lidar-based motion estimation, scan matching associates point-cloud data collected along a trajectory to a common reference frame, and computes an estimate for the trajectory. Most existing scan matching techniques are variants of the Iterative Closest Point (ICP) algorithm, which Pomerleau et al.~\cite{pomerleau2013comparing} provided a comparative study.

\begin{figure}[!htbp]
  \centering
  \includegraphics[height=2.2in]{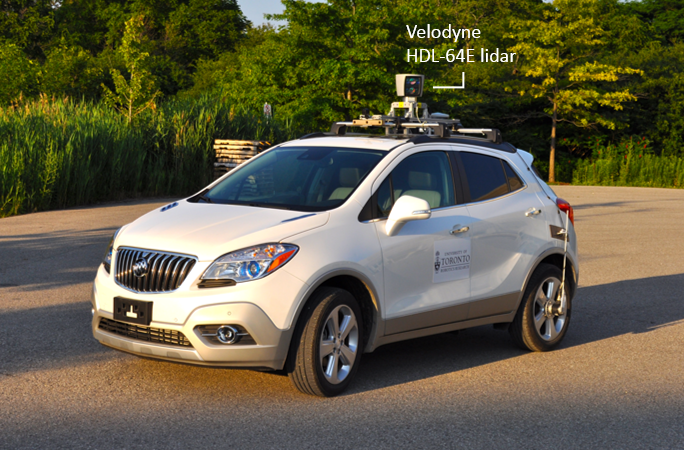}
  \caption{\footnotesize The Buick Encore test vehicle used for collecting data on University of Toronto campus. The vehicle is equipped with a Velodyne HDL-64E lidar, and an Applanix POS-LV system for ground truth.}
  \label{fig:car}
  \vspace{-5mm}
\end{figure}

The performance of odometry is crucial to tasks in autonomous navigation, such as mapping and localization. For lidar-based mapping, the performance of odometry directly influences the quality of the map, as drift in lidar odometry can cause the generated point-cloud map to be misaligned. For localization against the map, estimates from lidar odometry can act as an initial condition for localization. Furthermore, in a route-following system, the accuracy of lidar odometry dictates how well the robot can follow its path without a successful localization.

Biases in lidar-based motion estimation can result from a number of factors such as sensor noise, lidar beam divergence \cite{glennie2010static}, and the inherent bias in maximum-likelihood estimators. Moreover, we observe experimentally that biases in motion estimation are also correlated to the geometric distribution of points in a point-cloud.

\begin{figure*}[!htbp]
\centering
  \centering
  \adjincludegraphics[height=1in,trim={0cm 0.2cm 0cm 0.2cm},clip]{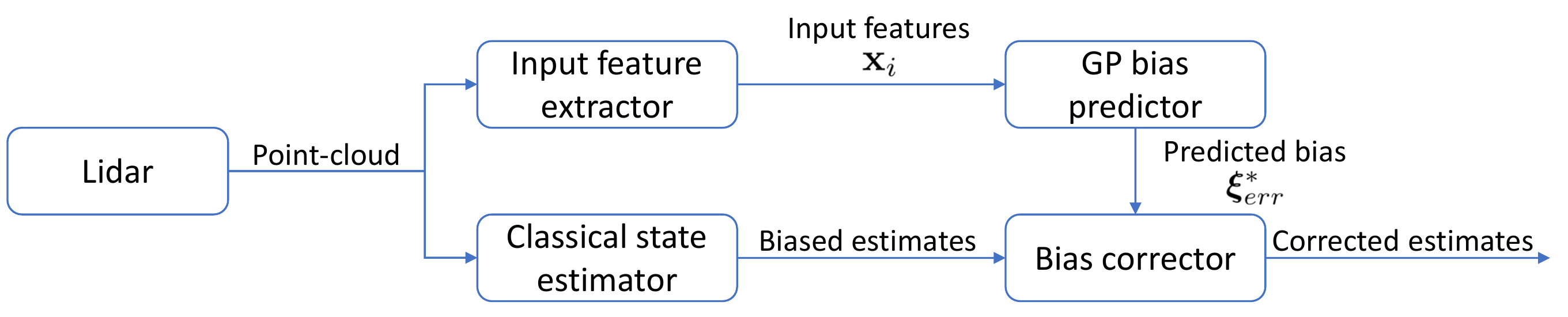}
\captionof{figure}{\footnotesize \label{fig:block} Biases are predicted from the GP model using inputs derived from point-cloud geometry. After estimates are computed by a classical state estimator, the predicted biases are applied as corrections to the estimates.}
\vspace{-5mm}
\end{figure*}

We present a novel technique to correct for biases in classical motion estimation using a machine learning approach. We fit a GP model using training data. The inputs to the GP model are high-level features derived from the geometric distribution of points in a point-cloud. The model predicts for the expected bias in the poses computed by a state estimator, which can be directly applied back as a correction to the state estimator. The prediction and correction step can be performed online with minimal computational effort. A high-level overview of the bias prediction and correction process is shown in the block diagram in Figure \ref{fig:block}.

Existing algorithms for lidar-based motion estimation differ in the types of scan matching algorithm used (point-to-plane ICP, point-to-point ICP, etc.), as well as the keypoint extraction strategy. We stress that regardless 
of how a motion estimation pipeline is formulated, there will evidently be residual pose errors left. Thus the estimates can always benefit from corrections made using a predicted error.

This work brings the following contributions:
\begin{itemize}
\item a novel technique to directly correct for the output of a classical state estimator using a learned bias correction
\item the demonstration of how scene geometry can be used to model biases in motion estimation
\end{itemize}

We demonstrate our technique through lidar-only motion estimation, but it can be easily extended to other motion estimation pipelines such as stereo visual odometry (VO). 

In Section \ref{sec:related_work} we review the previous work in related fields. A high-level overview of our lidar odometry pipeline is provided in Section \ref{sec:LO}. Section \ref{sec:error_prediction_and_correction} describes the details of predicting biases using GP regression and applying the bias correction to our classical estimator. Section \ref{sec:results} shows evaluation of our technique on the KITTI odometry dataset and a University of Toronto campus dataset. In Section \ref{sec:conclusion_and_future_work} we give concluding remarks and discuss future work.

\section{Related Work}
\label{sec:related_work}
Our method falls into the category of improving lidar-based motion estimation, which has been a well-explored field in the past decade. Segal et al. \cite{segal2009generalized} developed the Generalized ICP (GICP) by combining point-to-point ICP and point-to-plane ICP into a single framework. Serafin et al. \cite{serafin2015nicp} developed the Normal Iterative Closest Point (NICP) algorithm by considering both the normal and the local surface information for each point, and showed an overall improvement over GICP. Variants of the Normal Distribution Transforms (NDT) algorithms were developed as alternatives to ICP \cite{magnusson2007scan}, \cite{stoyanov2012point}, where the point-cloud is discretized and represented by a combination of normal distributions. Magnusson et al. \cite{magnusson2009evaluation}, \cite{magnusson2015beyond} compared NDT and ICP algorithms, and showed NDT is generally more accurate and may converge faster. State-of-the-art lidar estimation method, LOAM \cite{zhang2014loam}, achieves accurate and efficient scan matching by having two algorithms (odometry and mapping) running in parallel. Odometry runs at a higher frequency to estimate the velocity of the sensor and unwrap the motion-distorted point-clouds, while mapping runs at a lower frequency but with higher fidelity to cancel the drift in odometry.

Other than using improved scan matching algorithms, the performance of motion estimation can also be enhanced by choosing keypoints that are stable and provide sufficient constraints to the problem. LOAM selects keypoints on planes and edges. Sefarin et al. \cite{serafin2016fast} used segmentation to extract keypoints on lines and planes after removing ground points, and showed an improvement in the estimated trajectory than the commonly used NARF keypoints \cite{steder2011point}.

Previous work has been done to relate the geometry of points to the accuracy of state estimation solutions. Gelfand et al. \cite{gelfand2003geometrically} showed that in point-to-plane ICP, points with normals perpendicular to a direction provide no constraints to that direction. Similarly, Zhang et al. \cite{zhang2016degeneracy} showed the lack of geometric structures can lead to degeneracy, making the estimation problem ill-conditioned in certain directions. Our method differs from \cite{gelfand2003geometrically} and \cite{zhang2016degeneracy} in that rather than trying to mathematically quantify how the geometry of points can generate errors in odometry, we learn the error from training data using features derived from the geometry of points.
% Our method also has the advantage that it can discover other hidden internal biases in an estimator, since it uses training data produced by the same estimator.
 
Efforts have also be made on techniques to compensate for bias in motion estimation. The work in \cite{dubbelman2009bias} showed biases in stereo-vision-based motion estimation can arise due to incorrectly modelling the noise distribution of landmarks, and proposed a technique to compensate for the error. The work in \cite{farboud2014towards} also applies a bias correction to an estimator, but does not learn the correction from training data. Rather, by quantifying how bias increases with measurement noise, the work in \cite{farboud2014towards} is able to compute a corrected estimate for a hypothetical noise-free scenario. Peretroukhin et al. \cite{peretroukhin2017reducing} reduced drift in VO by using Convolutional Neural Networks (CNN) to infer sun direction. Most related to our work is the method proposed by Hidalgo-Carri\'{o} et al. \cite{hidalgo2017gaussian}, in which a GP model was used to predict for errors in wheel odometry, which is part of a SLAM system. Results in \cite{hidalgo2017gaussian} shows that by selecting image frames for VO adaptively based on the predicted errors in wheel odometry, the estimated trajectory did not lose significant accuracy while using much less image frames than selecting image frames non-adaptively. However, the work in \cite{hidalgo2017gaussian} is limited to adaptively distributing image frames using predicted wheel odometry errors, while our method directly applies a learned bias correction to the trajectory computed by a state estimator. Moreover, the work in \cite{hidalgo2017gaussian} did not use any publicly available datasets and was only evaluated on very short trajectories. On the contrary, we verify our method on data collected over more than 50\,km of traversal, including the publicly available KITTI odometry benchmark.

\section{Lidar Odometry Algorithm}
\label{sec:LO}
\subsection{Point-cloud downsampling}
Our odometry algorithm operates on keypoints. In other words, we downsample the point-cloud (or extract keypoints) prior to scan matching, rather than using the raw, dense point-clouds. We select keypoints based on normalized intensity values as well as whether a point is on a plane.

Assuming Lambertian reflectance, the intensity value of a point, $I,$  is inversely proportional to $r^2$ \cite{kashani2015review}, where $r$ is the range of the point. We define normalized intensity to be $Ir^2.$ To determine whether a point lies on a planar surface, similar to \cite{serafin2016fast}, we look at the eigenvalues of the covariance matrix of its $k$-nearest neighbours. This computation is performed using the open-source registration library \textit{libpointmatcher} \cite{pomerleau2013comparing}. The eigenvalues are sorted such that $\lambda_1$ is the smallest eigenvalue and $\lambda_3$ is the largest. If a point is on a planar surface, then $\lambda_1$ will be much smaller than $\lambda_2$ and $\lambda_3.$ 

A point may be selected as a keypoint if it satisfies either of the following two conditions:

\begin{itemize}
\item Its normalized intensity is greater than a threshold:
\begin{center}
$Ir^2 > \epsilon_I$
\end{center}
\item $\lambda_1$ is much smaller than $\lambda_2$ and $\lambda_3,$ which we denote using the following ratio between the eigenvalues:
\begin{center}
$\frac{\lambda_1 + \lambda_2 + \lambda_3}{\lambda_1} > \epsilon_p$
\end{center}
\end{itemize}

In our experiments, the thresholds $\epsilon_I$ and $\epsilon_p$ are chosen such that approximately 5\% of points are kept as keypoints and used for odometry. Ground points were ignored when selecting points on planar surfaces. Two conditions were set for selecting keypoints, so that even in environments lacking planar structures, a sufficient number of keypoints can still be selected using the first condition.

\subsection{Point Matching}
Given two downsampled point-clouds, point matches are selected based on Euclidean distance. Point matching is handled through the library \textit{libnabo} \cite{amigoni2009insightful}, which implements an efficient nearest-neighbour search with k-d trees.

\subsection{Trajectory Estimation}
For trajectory estimation we adopt the Simultaneous Trajectory Estimation and Mapping (STEAM) \cite{anderson2015full} framework, in which continuous-time trajectory estimation is carried out as GP regression. However, we do not save any landmark positions, as we are only solving for odometry rather than the full STEAM problem. Note that the GP regression problem for continuous-time trajectory estimation is different from the GP regression used for predicting odometry bias.

\par
We formulate two types of measurement cost terms. Let $\mathbf{p}$ be a point measured at time $t = k,$ and let $\mathbf{q}$ be its matched point expressed in the inertial frame. Define $\mathbf{e}_j = \mathbf{q} - \mathbf{T}_{0,k}\mathbf{p},$ where $\mathbf{T}_{0,k}$ is the queried pose for time $t = k.$ If $\mathbf{p}$ lies on a planar surface, then we formulate the point-to-plane whitened error norm:

\begin{equation}
\label{eq:point-to-plane}
u_j^{\textrm{plane}} = \sqrt{ \mathbf{e}_j^T(\mathbf{n}\mathbf{n}^T)\mathbf{e}_j}
\end{equation}
where $\mathbf{n}$ is the normal vector. If $\mathbf{p}$ does not lie on a plane, then we formulate the point-to-point whitened error norm:

\begin{equation}
\label{eq:point-to-point}
u_j^{\textrm{point}} = \sqrt{\mathbf{e}_j^T\mathbf{R}_j^{-1} \mathbf{e}_j}
\end{equation}
where $\mathbf{R}_j$ is the associated measurement covariance. Given the whitened error norm for a matched pair of points, we use the Geman-McClure robust cost function to build the corresponding measurement cost term:
\begin{equation}
\rho(u_j) = \frac{1}{2}\frac{u_j^2}{1+u_j^2}
\end{equation}
where $u_j$ is the whitened error norm. The full objective function which we seek to optimize is composed of cost terms associated with measurements and a constant-velocity motion prior \cite{anderson2015full}.

\par
It is worth noting that when forming the cost terms, for each point an interpolated pose is formulated given its time-stamp, $\mathbf{T}_{0,k}.$ This is in contrast with standard, discrete-time algorithms, in which all points in a point-cloud are assumed to share the same time-stamp and pose transform.

\par
The Velodyne data from the KITTI datasets were processed to eliminate the effect of motion-distortion, therefore all points in a point-cloud are treated as being measured at the same time. However, when working with raw lidar data, it is crucial to take into account that the sensor takes measurements continuously as it moves along a trajectory. We demonstrate this capability by running our odometry algorithm on a motion-distorted lidar dataset we collected around the University of Toronto campus.

\section{Error Prediction and Correction}
\label{sec:error_prediction_and_correction}
\subsection{Odometry Error Evaluation}
We define a frame as a full sweep ($360^\circ$) of the lidar, and we query our continuous-time trajectory for each frame. For a data sequence, this results in $N$ queried poses, $\mathbf{T}_{1,0}, \mathbf{T}_{2,0}, \hdots, \mathbf{T}_{N,0},$ where $N$ is the number of frames.

\par
To evaluate odometry errors we define a window of $\kappa$ frames. For each frame $\tau,$ where $\tau \geq \kappa,$ we calculate the relative pose change from frame $\tau-\kappa,$ and compare that against ground truth to compute an error:

\begin{equation}
\mathbf{T}_{\textrm{odom}_{\tau, \tau-\kappa}} = \mathbf{T}_{\textrm{odom}_{\tau,0}} \mathbf{T}_{\textrm{odom}_{\tau-\kappa,0}}^{-1}
\end{equation}

\begin{equation}
\mathbf{T}_{\textrm{gt}_{\tau, \tau-\kappa}} = \mathbf{T}_{\textrm{gt}_{\tau,0}} \mathbf{T}_{\textrm{gt}_{\tau-\kappa,0}}^{-1}
\end{equation}

\begin{equation}
\mathbf{T}_{\textrm{err}_{\tau, \tau-\kappa}} = \mathbf{T}_{\textrm{gt}_{\tau, \tau-\kappa}} \mathbf{T}_{\textrm{odom}_{\tau, \tau-\kappa}}^{-1}
\end{equation}
where $\mathbf{T}_{\textrm{odom}}$ is pose estimated by lidar odometry, $\mathbf{T}_{\textrm{gt}}$ is the ground truth pose, and $\mathbf{T}_{\textrm{err}}$ is the odometry error. We evaluate the odometry error for $\tau = \kappa, \kappa+1, \hdots, N.$ 

\par
Rather than always evaluating pose change from the previous frame, $\mathbf{T}_{\tau, \tau-1},$ we evaluate $\mathbf{T}_{\tau, \tau-\kappa}$ and leave $\kappa$ as a parameter of choice. This is due to the high frame rate of the sensor. The Velodyne lidar spins at 10\,Hz, therefore $\mathbf{T}_{\tau, \tau-1}$ is the pose change of the sensor over 0.1\,seconds, corresponding to a very short section of the trajectory. The imprecision in GPS measurements might make ground truth over such a short trajectory section noisy.

\par
For the error to be a valid output of a GP model, we convert $\mathbf{T}_{\textrm{err}} \in SE(3)$ to a vectorspace representation:

\begin{equation}
\bm{\xi}_{\textrm{err}_{\tau, \tau-\kappa}} = \mathrm{ln}(\mathbf{T}_{\textrm{err}_{\tau, \tau-\kappa}})^{\vee}
\end{equation}
where $\bm{\xi}_{\textrm{err}_{\tau, \tau-\kappa}} \in \mathbb{R}^{6}$ is the vectorspace representation of the error. The operator $\vee$ converts a $4\times 4$ member of the \textit{Lie algebra}, $\mathfrak{se}(3),$ to $\bm{\xi}= \begin{bmatrix}
\bm{\rho}^T & \bm{\phi}^T
\end{bmatrix}^T \in \mathbb{R}^6$ \cite{barfoot2014associating}, \cite{barfoot2017state}, where $\bm{\rho} = \begin{bmatrix}
\rho_1 & \rho_2 & \rho_3
\end{bmatrix}^T,$ and $\bm{\phi} = \begin{bmatrix}
\phi_1 & \phi_2 & \phi_3
\end{bmatrix}^T.$ We can convert $\bm{\xi}_{\textrm{err}_{\tau, \tau-\kappa}}$ back to $SE(3)$ via the exponential map:

\begin{equation}
\label{eq:ex_map}
\mathbf{T}_{\textrm{err}_{\tau, \tau-\kappa}} = \exp({\bm{\xi}_{\textrm{err}_{\tau, \tau-\kappa}}} ^\wedge)
\end{equation}
where $\wedge$ converts $\bm{\xi} \in \mathbb{R}^6$ to a member of $\mathfrak{se}(3)$ \cite{barfoot2014associating}, \cite{barfoot2017state}.

\subsection{Gaussian Process Model}
GP models offer a principled approach for learning from noisy observations. Gaussian processes has been widely used in robotics, such as in terrain assessment \cite{berczi2015learning}, building occupancy grid maps \cite{o2012gaussian}, and trajectory estimation \cite{anderson2015full}.

\par
We wish to predict the odometry error, $\bm{\xi}_{\textrm{err}},$ which is a $6\times1$ vector. While there are methods to handle GP regression with vector outputs, for simplicity we model each degree of freedom (DOF) of $\bm{\xi}_{\textrm{err}}$ separately. In other words, we fit a separate model for each element of the 6-DOF error vector.

\par
For $n$ observations in a dataset, we have $n$ $D$-dimensional inputs $\mathbf{X} = \begin{bmatrix}
\mathbf{x}_1 & \mathbf{x}_2 & \dots & \mathbf{x}_n
\end{bmatrix}^T,$ where $\mathbf{x}_i \in \mathbb{R}^{D},$ and $\mathbf{X}\in \mathbb{R}^{n\times D}.$ We also have $n$ scalar outputs $\mathbf{y} = \begin{bmatrix}
y_1 & y_2 & \dots & y_n
\end{bmatrix}^T,$ where $\mathbf{y} \in \mathbb{R}^n.$ Let $f$ to be the underlying relation between input and output, but the observations are noisy. Therefore we have $y_i = f(\mathbf{x}_i) + \epsilon,$ where $\epsilon$ is the noise, which we assume to be Gaussian with variance $\sigma_n^2.$ In our problem, the inputs $\mathbf{x}_i$ are features derived from geometry of the point-clouds. The scalar output $y_i$ is an element of the error vector $\bm{\xi}_{err},$ as shown in Section \ref{sec:input}.

%Given a dataset of $n$ observations, $\mathcal{D} = \{(\mathbf{x}_i, y_i)|i = 1, \dots, n\}.$ Assuming the inputs $\mathbf{x}_i$ are $D$-dimensional, then we can combine all inputs into a matrix $\mathbf{X} = \begin{bmatrix}
%\mathbf{x}_1 & \mathbf{x}_2 & \dots & \mathbf{x}_n
%\end{bmatrix}^T,$ where $\mathbf{x}_i \in \mathbb{R}^{D},$ and $\mathbf{X}\in \mathbb{R}^{n\times D}.$ Similarly, we can combine the scalar outputs $y_i$ into a vector $\mathbf{y} = \begin{bmatrix}
%y_1 & y_2 & \dots & y_n
%\end{bmatrix}^T,$ where $\mathbf{y} \in \mathbb{R}^n.$ Let $f$ to be the underlying relation between the input and the output, but the observations are noisy. Therefore we have $y_i = f(\mathbf{x}_i) + \epsilon,$ where $\epsilon$ is the noise. We assume $\epsilon$ to be Gaussian with variance $\sigma_n^2.$

\par
Given $\mathbf{X}$ and $\mathbf{y},$ to make predictions $\mathbf{f}^*$ on new data $\mathbf{X}^*,$ the predictive distribution for Gaussian process regression \cite{rasmussen2006gaussian} can be formulated as :
\begin{equation}
p(\mathbf{f}^* | \mathbf{X}, \mathbf{y}, \mathbf{X}^*) \sim \mathcal{N}(\mathbf{\bar{f}}^*, \mathrm{cov}(\mathbf{f}^*))
\end{equation}

\begin{equation}
\label{eqn:pred_mean}
\mathbf{\bar{f}^*} = \mathbf{K}(\mathbf{X}^*,\mathbf{X})[(\mathbf{K}(\mathbf{X},\mathbf{X})+(\sigma_n)^2\mathbf{I})]^{-1}\mathbf{y}
\end{equation}

\begin{dmath}
\mathrm{cov}(\mathbf{f^*}) = \mathbf{K}(\mathbf{X}^*,\mathbf{X}^*)-\mathbf{K}(\mathbf{X}^*,\mathbf{X})[(\mathbf{K}(\mathbf{X},\mathbf{X})+(\sigma_n)^2\mathbf{I})]^{-1}\mathbf{K}(\mathbf{X},\mathbf{X}^*)
\end{dmath}
where $\mathbf{\bar{f}}^*$ is the mean prediction and $\mathrm{cov}(\mathbf{f^*})$ is the variance. $\mathbf{K}(\mathbf{X}, \mathbf{X}) \in \mathbb{R}^{n \times n}$ is the kernel matrix with $\mathbf{K}_{ij} = k(\mathbf{x}_i, \mathbf{x}_j),$ where $k(\cdot, \cdot)$ is the kernel function. We use the squared exponential kernel function with a separate length scale for each input dimension \cite{rasmussen2006gaussian}:

\begin{equation}
k(\mathbf{x}_i, \mathbf{x}_j) = \sigma_f^2\exp(\frac{1}{2}(\mathbf{x}_i-\mathbf{x}_j)^T\mathbf{M}(\mathbf{x}_i-\mathbf{x}_j))
\end{equation}
where $\mathbf{M} \in \mathbb{R}^{D\times D}$ is a diagonal matrix with entries $l_1^{-2}, \dots, l_D^{-2},$ and $l_1, \dots, l_D$ are the characteristic length scales for each of the $D$ input dimensions. $\sigma_f$ is the signal standard deviation, and $\sigma_n$ is the noise standard deviation.

\par
$\Theta = \{l_1, \dots, l_D, \sigma_f, \sigma_n\}$ form the set of hyperparameters for the GP model. Define the log marginal likelihood \cite{rasmussen2006gaussian}:

\begin{equation}
\label{eq:ml}
\ln \ p(\mathbf{y}|\mathbf{X},\Theta) = -\frac{1}{2}\mathbf{y}^T\mathbf{K}_{\mathbf{y}}^{-1}\mathbf{y}-\frac{1}{2}\ln|\mathbf{K}_{\mathbf{y}}|-\frac{n}{2}\ln 2\pi
\end{equation}
where $\mathbf{K}_{\mathbf{y}} = \mathbf{K}(\mathbf{X},\mathbf{X})+(\sigma_n)^2\mathbf{I}.$ In practice the hyperparamters $\Theta$ are chosen by maximizing the log marginal likelihood in \eqref{eq:ml} with respect to  $\Theta.$

\subsection{Input Features}
\label{sec:input}
Choosing correct inputs for the GP model is crucial to its predictive capabilities. However, this is a non-trivial task if the output of the model is error in odometry. The method in \cite{hidalgo2017gaussian} used orientation angles, speed and position of joints, measurements of gyroscopes, and IMU measurements as inputs to model errors in wheel odometry. Selecting inputs is less obvious in our situation, however, as we do not have measurements from any sensors other than the lidar. In our method, the inputs are selected based on high-level features derived from the geometry of points in a point-cloud. We show that we can achieve significant reduction in odometry error, using only inputs derived from the point-clouds and no other measurements.

\par
\begin{figure}[!htbp]
  \centering
  \includegraphics[height=1.75in]{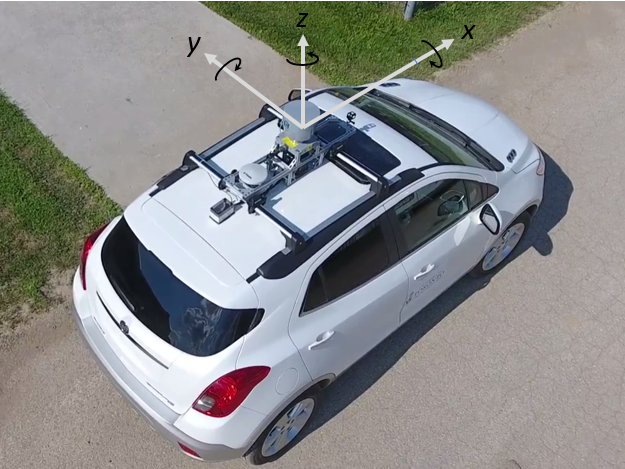}
  \caption{\footnotesize The coordinate system used by our odometry algorithm. Roll, pitch, and yaw are rotations about the $x,$ $y,$ and $z$ axes, respectively.}
  \vspace{-5mm}
  \label{fig:coord}
\end{figure}

We fit a GP model to each DOF of odometry error. This requires choosing a set of input features for each element of the error vector $\bm{\xi}_{err}$ we wish to model. The majority of odometry errors for our odometry algorithm are in the directions of $z\,(\rho_3),$ pitch$\,(\phi_2
),$ and roll$\,(\phi_1
),$ while our algorithm is relatively accurate in the directions of $x\,(\rho_1),$ $y\,(\rho_2),$ and yaw$\,(\phi_3
)$ (see Figure \ref{fig:coord} for the coordinate system our odometry algorithm uses). Therefore, we only make predictions and corrections in 3-DOF for $z,$ pitch, and roll.

\subsubsection{z and pitch}
We use the same input to predict for errors in $z$ and pitch. For $z,$ the output of the GP model is:
\begin{equation}
e_{\rho_3} = \begin{bmatrix}
0 & 0 & 1 & 0 & 0 & 0
\end{bmatrix} \bm{\xi}_{\textrm{err}}
\end{equation}
and for pitch:
\begin{equation}
e_{\phi_2} = \begin{bmatrix}
0 & 0 & 0 & 0 & 1 & 0
\end{bmatrix} \bm{\xi}_{\textrm{err}}
\end{equation}

Our choices for candidate input features are inspired by the work in \cite{gelfand2003geometrically}, where the authors argue that in an ICP problem, the distribution of normal vectors affects how well-constrained the solution is in each degree of freedom. A number of input features based on the distribution of surface normals were tested, and we select input features based on evaluation on the training sequences of the KITTI benchmark, as described in Section \ref{sec:training}. Here we present our final choice for input features, which resulted in the greatest reduction in odometry errors after applying correction on the KITTI training sequences, among all input features tested.

\par
For the $3\times1$ normal vector $\mathbf{n} = \begin{bmatrix}
n_x & n_y & n_z
\end{bmatrix}^T,$ we sum each component of the normal over all points lying on planar surfaces to form our 3-dimensional input feature, and normalize by the number of points. Points not on planar surfaces are ignored, since their normal estimates are noisy. If the (downsampled) point-cloud associated with frame $i$ has $M$ points, in which $P$ of them are on planar surfaces, then the input can be calculated as:

\begin{equation}
\mathbf{x}_i = \frac{1}{M}\begin{bmatrix}
\sum\limits_{p=1}^{P} \norm{n_{p,x}} & \sum\limits_{p=1}^{P} \norm{n_{p,y}} & \sum\limits_{p=1}^{P} \norm{n_{p,z}}
\end{bmatrix}^T
\end{equation}
where $\mathbf{n} = \begin{bmatrix}
n_{p,x} & n_{p,y} & n_{p,z}
\end{bmatrix}^T$ is the normal for point $p.$

\subsubsection{roll}
\begin{figure}[!htbp]
  \centering
  \includegraphics[height=1.5in, trim={1cm 0.5cm 1cm 2cm}]{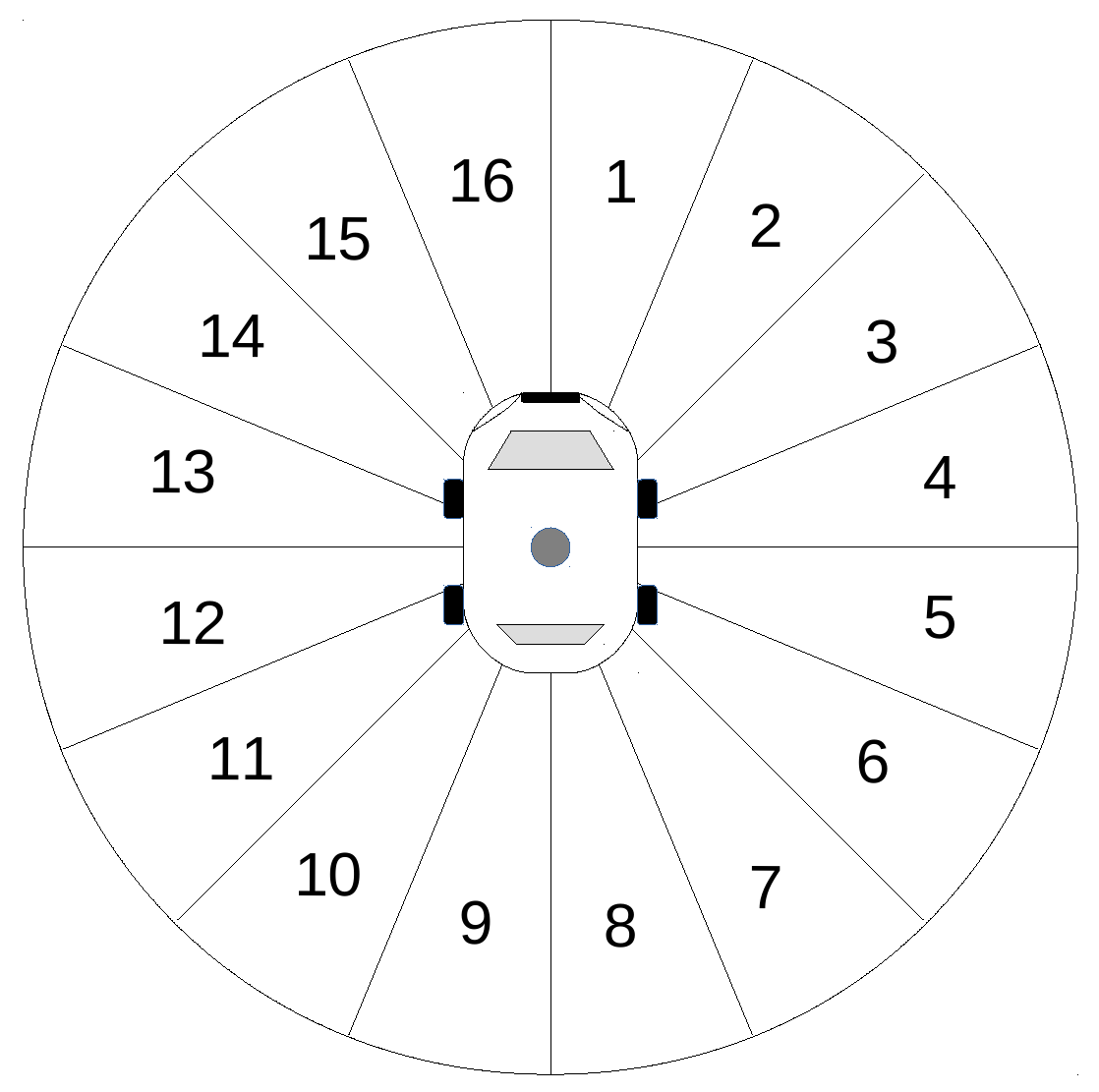}
  \caption{\footnotesize We divide the point-cloud into 16 ``slices'' based on azimuth. For each ``slice'', we calculate the number of points with normals in the $z$-direction and normalize. This forms our 16-dimensional input $\mathbf{x}_i$ for roll.}

  \label{fig:pizza}
\end{figure}

We discretize the point-cloud into 16 equally spaced ``slices'' based on azimuth, as shown in Figure \ref{fig:pizza}. A drawing of a car is included for reference. For each ``slice'', we calculate the number of points on planar surfaces with normal vector pointing in the $z$-direction, and normalize by the total number of points. This forms our 16-dimensional input. The output of the GP model for roll is:

\begin{equation}
e_{\phi_1} = \begin{bmatrix}
0 & 0 & 0 & 1 & 0 & 0
\end{bmatrix} \bm{\xi}_{\textrm{err}}
\end{equation}

\vspace{-5mm}
\subsection{Applying Correction to Odometry}
Given odometry estimates from new data, for each frame $\tau \geq \kappa,$ we predict for the odometry error between frames $\tau$ and $\tau - \kappa,$ which we denote by $\bm{\xi}_{\textrm{err}_{\tau, \tau-\kappa}}^*.$ We then apply the predicted errors as corrections to the estimates, by converting the predicted error from vectorspace back to $SE(3)$ using \eqref{eq:ex_map}. We assume the error is accumulated uniformly from frame $\tau-\kappa$ to $\tau,$ as shown in Algorithm \ref{algo:correction}. The prediction and correction step is applied to the poses for frames $\tau = \kappa, \kappa+1, \dots, N,$ where $N$ is the total number of frames. 

\begin{algorithm}
\caption{Applying correction to odometry}\label{algo:correction}
\begin{algorithmic}[1]
\State Let $\mathbf{T}_{\textrm{corr}_{\kappa-1, 0}} = \mathbf{T}_{\textrm{odom}_{\kappa-1, 0}}$
\For{$\tau = \kappa, \kappa+1, \dots, N$}
\State predict for odometry error $\bm{\xi}_{\textrm{err}_{\tau, \tau-\kappa}}^*$
\State $\delta\bm{\xi}_{\textrm{err}}^* = \frac{1}{\kappa}\bm{\xi}_{\textrm{err}_{\tau, \tau-\kappa}}^*$
\State $\delta\mathbf{T}_{\textrm{err}}^* = \exp({\delta\bm{\xi}_{\textrm{err}}^*}^{\wedge})$
\State $\mathbf{T}_{\textrm{odom}_{\tau, \tau-1}} = \mathbf{T}_{\textrm{odom}_{\tau,0}}\mathbf{T}_{\textrm{odom}_{\tau-1,0}}^{-1}$
\State $\mathbf{T}_{\textrm{corr}_{\tau, \tau-1}} = \delta\mathbf{T}_{\textrm{err}}^* \mathbf{T}_{\textrm{odom}_{\tau, \tau-1}}$
\State $\mathbf{T}_{\textrm{corr}_{\tau, 0}} = \mathbf{T}_{\textrm{corr}_{\tau, \tau-1}} \mathbf{T}_{\textrm{corr}_{\tau-1, 0}}$
\EndFor

\end{algorithmic}
\end{algorithm}

Shown in Algorithm \ref{algo:correction}, $\mathbf{T}_{\textrm{odom}_{\tau,0}}$ is the pose for frame $\tau$ before applying the correction, and $\mathbf{T}_{\textrm{corr}_{\tau, 0}}$ is the pose for frame $\tau$ after correction is applied. In practice, $\kappa$ is set between 2 to 20. Since we only make predictions in $z$, pitch, and roll, we have $\bm{\xi}^*_{err} = \begin{bmatrix}
0&0&e_{\rho3}^*&e_{\phi1}^*&e_{\phi2}^*&0
\end{bmatrix}^T.$

\par
To evaluate the mean prediction as in \eqref{eqn:pred_mean}, the latter term $[(\mathbf{K}(\mathbf{X},\mathbf{X})+(\sigma_n)^2\mathbf{I})]^{-1}\mathbf{y}$ only needs to be determined once, which requires minimal computational effort.

\section{Results}
\label{sec:results}
\subsection{Evaluating on KITTI Training Sequences}
\label{sec:training}

\begin{figure*}[!htbp]
\centering
\begin{minipage}{.5\textwidth}
  \centering
  \adjincludegraphics[height=1.7in,trim={2cm 1.9cm 2cm 3.3cm},clip]{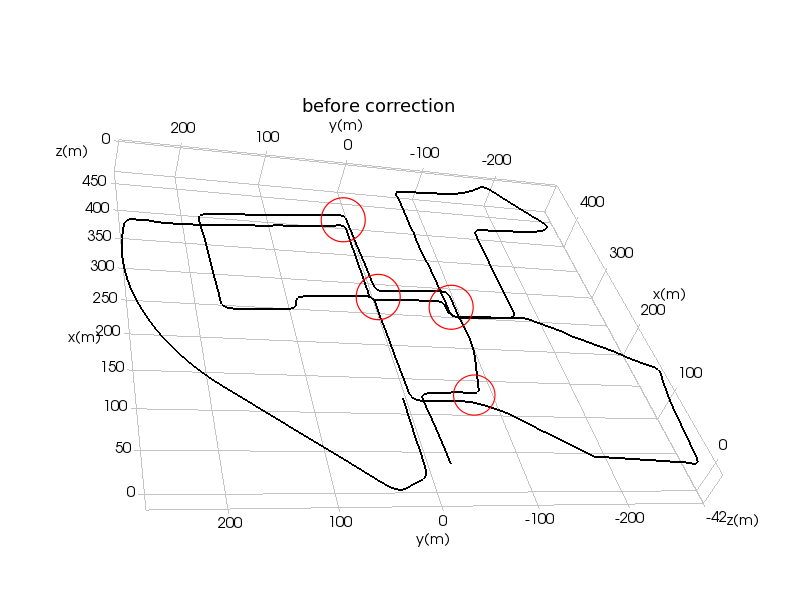}
  \label{fig:0_before}
\end{minipage}%
\begin{minipage}{.5\textwidth}
  \centering
    \adjincludegraphics[height=1.7in,trim={2cm 1.9cm 2cm 3.3cm},clip]{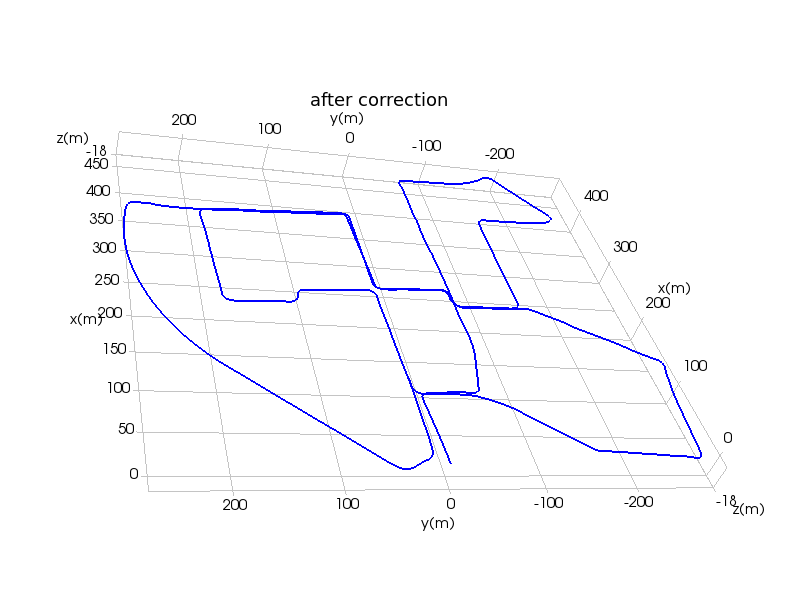}
  \label{fig:0_after}
\end{minipage}
\captionof{figure}{\footnotesize \label{fig:0} 3D plots of odometry estimates for sequence 0 from the same perspective. Black is odometry before correction and blue is odometry after correction. Left: due to errors in roll, when the vehicle travels back to a location it visited before, the odometry aligns poorly (circled). Right: after applying the correction, when the vehicle travels back to the same location, the odometry estimate overlaps well with the previous trajectory segment.}
\end{figure*}

In Gaussian process regression, model selection is the process of making choices about the details of the model. For our problem, model selection involves choosing the input to the model, as well as setting the hyperparamters.

\par
We use the first 11 sequences of the KITTI dataset for selecting the best input features since ground truth is available. To evaluate how well a specific set of inputs predict the odometry error, we use cross-validation on the training data. Specifically, we leave 1 sequence out as the validation sequence, and fit the model on the other 10 sequences. The fitted model is used to make predictions on the unseen validation sequence, and the predictions are used to correct for the odometry estimates of the validation sequence. We cross-validate by repeating this process for every sequence, such that we have the corrected odometry estimates for all 11 sequences. The KITTI benchmark evaluates percentage errors across path segments of lengths $100, 200, \dots, 800$ meters, and an average over all path segments is computed. A total error averaged over path segments evaluated for all 11 sequences is also reported. The total error before and after the correction are used to quantify the improvements from applying the learned bias correction.

\par The training is done offline using the Gaussian Process Regression tool of the MATLAB Statistics and Machine Learning Toolbox \cite{matlab}. To set the hyperparameters, the log marginal likelihood in \eqref{eq:ml} is maximized with respect to the hyperparameters using gradient-based optimization.

\par
We have experimented with a number of input features for predicting odometry errors in $z,$ pitch, and roll. The set of input features resulting in the greatest reduction in total odometry errors were selected, namely the input features shown in Section \ref{sec:input}. Using these input features, the odometry error before and after the correction on the first 11 sequences of KITTI are shown in Table \ref{tab:kitti_train}. The corrections improved the odometry for 8 of the 11 sequences. Over all 11 sequences, the error is reduced from $1.13\%$ to $1.03\%,$ accounting for a $9\%$ reduction. The prediction and correction steps  cost only 10\,s of computational time, which is approximately $0.5\%$ of the cost for computing odometry.

\begin{figure}[!htbp]

  \centering
  \includegraphics[height=1.5in,trim={1.5cm 1cm 1cm 4.5cm},clip]{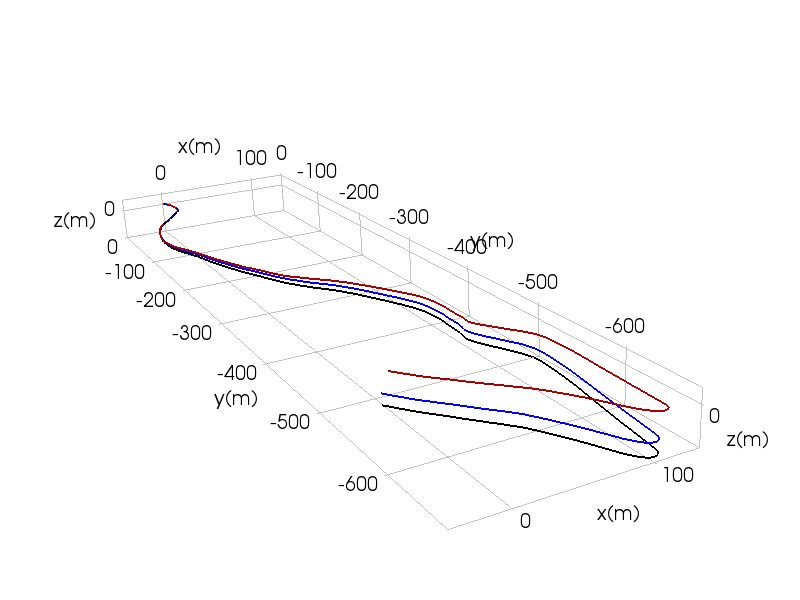}
  
  \caption{\footnotesize 3D plots of odometry estimates for sequence 10: uncorrected odometry estimates (black) vs. corrected odometry estimates (blue) when compared against ground truth (red). The corrections brought noticeable improvements over $z$ and pitch.}
  \label{fig:10}
\end{figure}

\begin{table}[]
\centering
\label{my-label}
\caption{\footnotesize \label{tab:kitti_train} Odometry errors before and after the correction for sequences 0 to 10 of the KITTI dataset.}

\begin{tabular}{cccc}
\begin{tabular}[c]{@{}c@{}}Sequence \\ no.\end{tabular} & \begin{tabular}[c]{@{}c@{}}Number \\ of frames\end{tabular} & \begin{tabular}[c]{@{}c@{}}Uncorrected \\ odometry error (\%)\end{tabular} & \begin{tabular}[c]{@{}c@{}}Corrected \\ odometry error(\%)\end{tabular} \\ \hline
0                                                       & 4541                                                        & 1.5465                                                                     & \textbf{1.427}                                                          \\
1                                                       & 1101                                                        & 2.2232                                                                     & \textbf{1.9627}                                                         \\
2                                                       & 4661                                                        & 0.9862                                                                     & \textbf{0.8613}                                                         \\
3                                                       & 801                                                         & \textbf{0.7296}                                                            & 0.8381                                                                  \\
4                                                       & 271                                                         & 0.6206                                                                     & \textbf{0.4802}                                                         \\
5                                                       & 2761                                                        & 0.6064                                                                     & \textbf{0.5316}                                                         \\
6                                                       & 1101                                                        & 0.5013                                                                     & \textbf{0.4106}                                                         \\
7                                                       & 1101                                                        & \textbf{0.6795}                                                            & 0.7579                                                                  \\
8                                                       & 4071                                                        & 1.0585                                                                     & \textbf{1.0007}                                                         \\
9                                                       & 1591                                                        & \textbf{0.9478}                                                            & 1.0023                                                                  \\
10                                                      & 1201                                                        & 1.7572                                                                     & \textbf{1.3501}                                                         \\
Total                                                 &                                                             & 1.1288                                                                     & \textbf{1.0294}                                                        
\end{tabular}
  \vspace{-5mm}
\end{table}

\par
Figure \ref{fig:10} shows a sequence where the uncorrected odometry results (black) are biased in $z$ and pitch. After applying corrections, the estimates (blue) are noticeably more aligned with ground truth (red). Figure \ref{fig:0} shows the results for sequence 0, where the odometry estimate does not overlap with itself when the vehicle travels back to somewhere it has been before, mostly due to errors in roll. However, this effect is greatly mitigated after the correction is applied.

\begin{figure*}[!htbp]
\centering
\begin{minipage}{.5\textwidth}
  \centering
  \adjincludegraphics[height=1.7in,trim={0cm 4cm 0cm 6cm},clip]{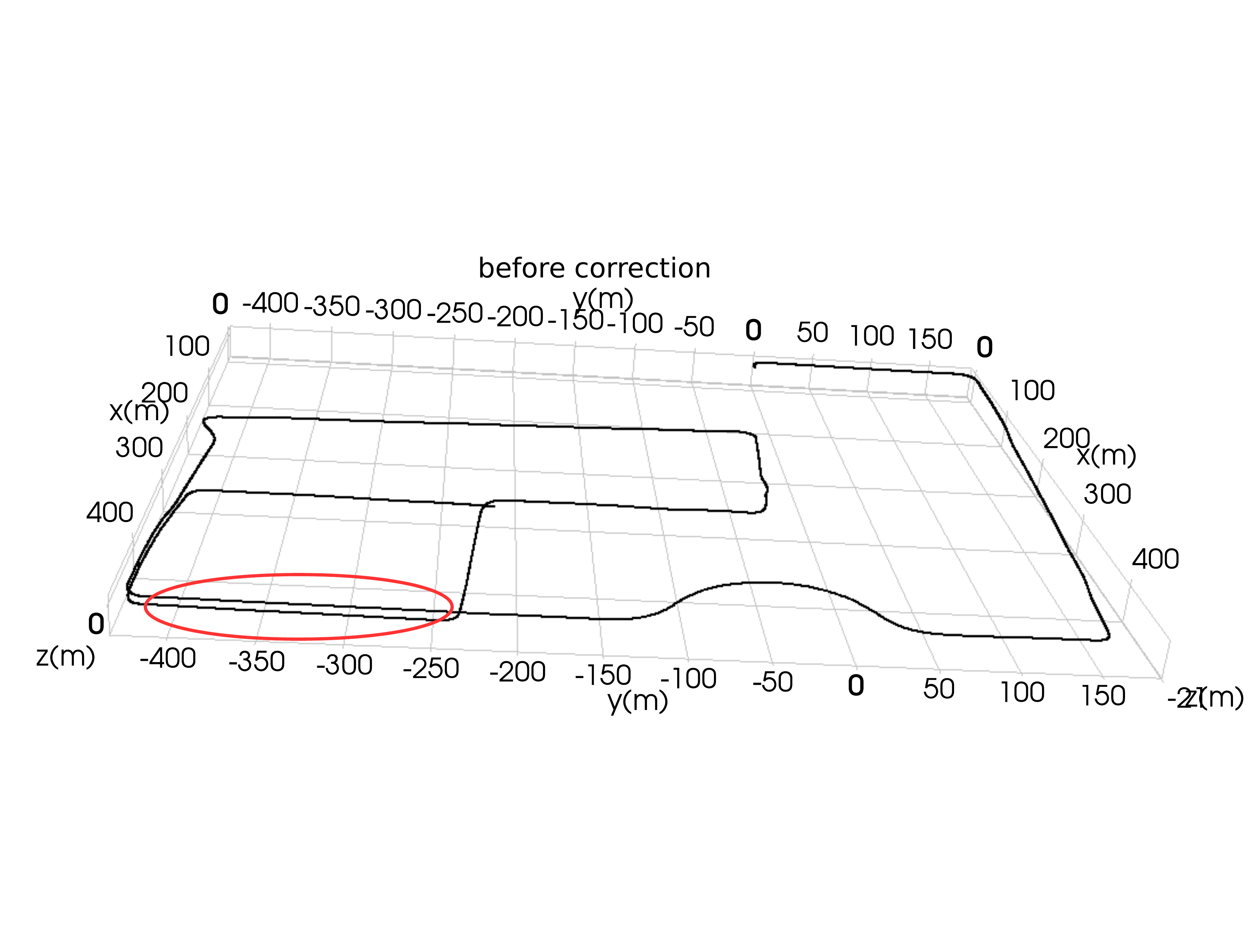}
  \label{fig:george4_before}
\end{minipage}%
\begin{minipage}{.5\textwidth}
  \centering
    \adjincludegraphics[height=1.7in,trim={0cm 4cm 0cm 6cm},clip]{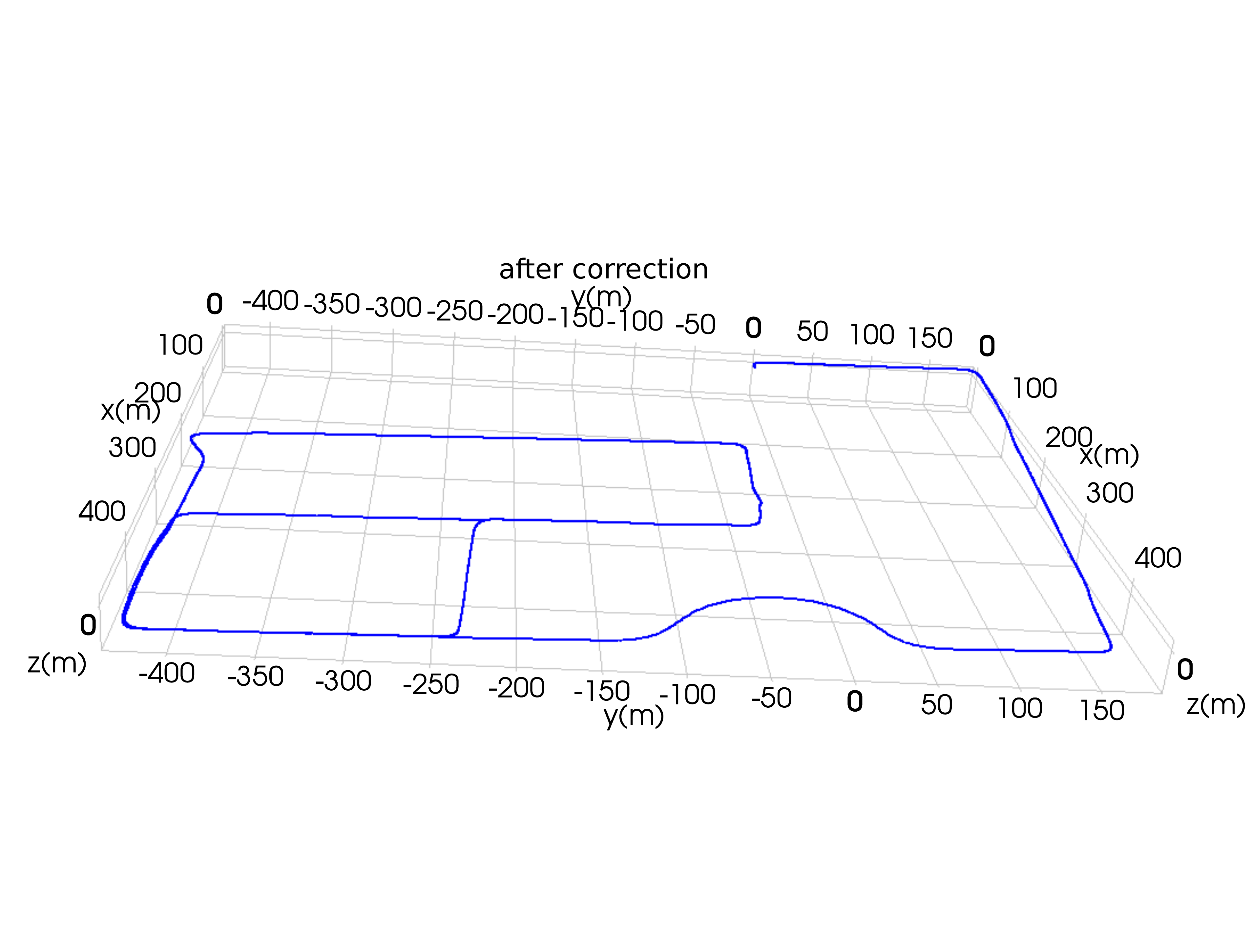}
  \label{fig:george4_after}
\end{minipage}
\vspace{-5mm}
\captionof{figure}{\footnotesize \label{fig:george4} 3D plots of odometry estimates for sequence 2 of the University of Toronto campus datasets from the same perspective. Black is odometry before correction and blue is odometry after correction. Left: due to errors, odometry does not overlap when the vehicle travels back to a path it has been before (circled). Right: after applying the correction, the odometry estimate overlaps well when the vehicle travels back to the same path.}
\vspace{-5mm}
\end{figure*}

\begin{figure}[!htbp]
  \centering
  \includegraphics[height=2in]{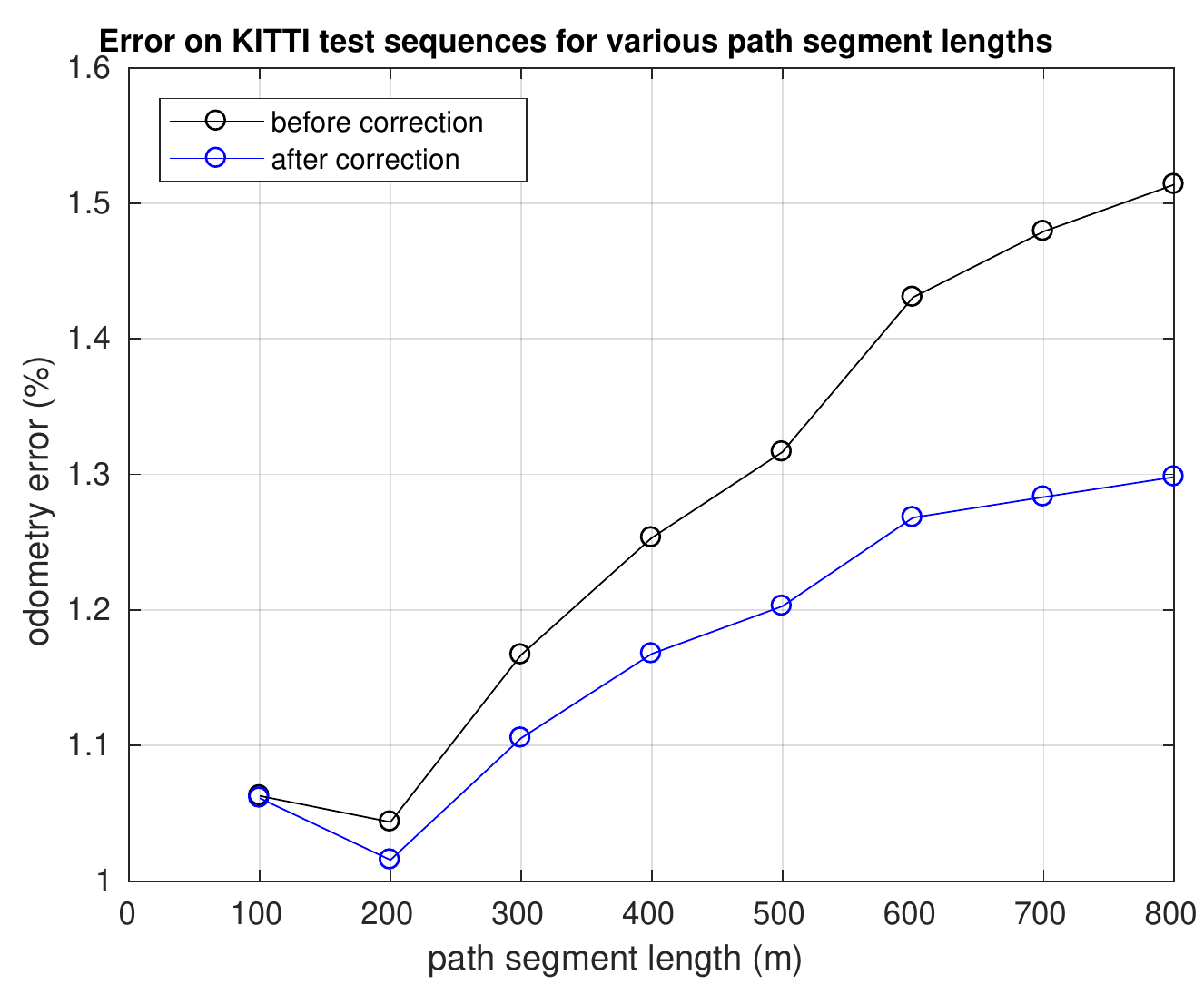}
  \caption{\footnotesize Error before and after correction on the KITTI test sequences. Our method produced improvements for path segments of all lengths between 100 to 800\,m. Specifically, for path segments of 800m, the odometry error decreased from 1.51\% to 1.30\%, resulting in a 14\% improvement.}
  \label{fig:test}
  \vspace{-5mm}
\end{figure}

\subsection{Results on KITTI Test Sequences}
Sequences 11-21 are the test sequences of KITTI, and ground truth is unavailable for these sequences. For evaluating against the 11 test sequences, we fit a model using all training sequences 0-10 with input features described in Section \ref{sec:input}. The predicted error is applied as a correction to each of sequences 11-21. Figure \ref{fig:test} shows the odometry error over all test sequences before and after the correction. Our method showed significant reduction in odometry error for path segments of all lengths from 100m to 800m. The improvement is more pronounced the longer the path segment. For path segments of 800m, the error is reduced from 1.51\% to 1.30\%, equivalent to a 14\% reduction.

\par
The total error over all path segments for all test sequences was reduced from 1.26\% down to 1.16\%, accounting for an 8\% reduction. Our odometry algorithm is accurate even before applying any corrections, ranking 3rd on the KITTI leader board at the time of submission among methods that use lidar only. Our uncorrected result currently ranks 4th among lidar-only methods as our corrected result now ranks 3rd. In fact, the top 2 methods for lidar only algorithm either performs SLAM (IMLS-SLAM), or has a higher-fidelity algorithm running in parallel to cancel the drift in odometry (LOAM). In contrast, our algorithm is strictly only odometry, and does not need loop closures or a second estimation algorithm to reduce the drift. By predicting errors using GP regression and applying them back as a bias correction, we achieved significant performance improvements over an already accurate odometry algorithm.

\subsection{Results on University of Toronto campus dataset}
A dataset was collected by our test vehicle (Figure \ref{fig:car}) along different routes around University of Toronto. This resulted in 7 sequences of Velodyne data over 18\,km of traversal. Similar to the KITTI dataset, 6-DOF ground truth is also available for the University of Toronto campus dataset. For consistency, to evaluate for odometry errors we use the same method as the KITTI benchmark, where errors are evaluated across path segments of lengths $100, 200, \dots, 800$ meters. We do not post-process the point-clouds to eliminate motion-distortion, but rather rely on our continuous-time odometry pipeline to address motion-distortion. 

\par
First, we attempted to fit a model using the training sequences of KITTI, and predict for odometry corrections on the University of Toronto campus dataset. However, this did not improve the odometry due to the considerable differences between the two datasets, including different calibration parameters for the Velodyne sensor, different systems for obtaining ground truth, and whether the lidar data is post-processed. Rather, 
we use cross-validation among sequences of the University of Toronto campus dataset to demonstrate the effectiveness of our method on this dataset. For each sequence, we leave it out as validation sequence and fit a model on the other 6 sequences, and the fitted model is used to make predictions on the unseen validation sequence. We do, however, use the same input features as selected by evaluating on the KITTI training sequences (Section \ref{sec:input}), as opposed to choosing another set of input features specifically for the University of Toronto campus dataset.

\par
The odometry errors before and after the correction are shown in Table \ref{tab:george}. The corrections improved 6 out of the 7 sequences, while the total error is reduced from $1.81\%$ to $1.56\%,$ accounting for a $14\%$ reduction in odometry errors. 
Figure \ref{fig:george4} shows plots of the odometry estimates for sequence 2 before and after the correction is applied. Before the correction, the estimated path does not overlap with itself when the vehicle travelled back to a street it has been before. This bias is eliminated after the correction is applied.

\begin{table}[]
\centering
\label{my-label}
\caption{\footnotesize \label{tab:george} Odometry errors before and after the correction for the University of Toronto campus dataset}

\begin{tabular}{cccc}
\begin{tabular}[c]{@{}c@{}}Sequence \\ no.\end{tabular} & \begin{tabular}[c]{@{}c@{}}Number \\ of frames\end{tabular} & \begin{tabular}[c]{@{}c@{}}Uncorrected \\ odometry error (\%)\end{tabular} & \begin{tabular}[c]{@{}c@{}}Corrected \\ odometry error(\%)\end{tabular} \\ \hline
0                                                       & 5000 & 1.6748 & \textbf{1.2696}                                                          \\
1                                                       & 5000 & 1.9166 & \textbf{1.3305}                                                         \\
2                                                     & 6000 & 1.6252 & \textbf{1.2102}                                                         \\
3                                                       & 6000                                                         & 2.2462 & \textbf{1.9529}                                                                                                                          \\
4                                                      & 3000                                                         & 2.1262 & \textbf{1.9334}                                                         \\
5                                                      & 6600                                                        & \textbf{1.5629} & 1.7260 \\
6                                                      & 7650                                                        & 1.7962 & \textbf{1.6395}                                                         \\

Total                                                 &                                                             & 1.8146 & \textbf{1.5598}                                                        
\end{tabular}
\vspace{-5mm}
\end{table}

\section{Conclusion and Future Work}
\label{sec:conclusion_and_future_work}
In this paper, we present a novel technique to directly correct for biases in a classical state estimator using a machine learning approach. A GP model is trained which takes features derived from scene geometry as inputs, and outputs a predicted bias, which is directly applied as a correction to the trajectory computed by the state estimator. Our method is demonstrated on lidar-only motion estimation, but can be easily generalized to other state estimation methods.

\par
The effectiveness of our technique is verified on the publicly available KITTI odometry dataset, and Velodyne data collected along different routes around the University of Toronto campus. Using the same hand-picked input features, our technique resulted in significant overall improvements to lidar odometry for all datasets tested. The next steps in extending this concept are to use more advanced models such as CNN, so that the input features are learned by the learning algorithm rather than hand-picked.

\section*{ACKNOWLEDGMENT}
We would like to thank Applanix Corporation and the Natural Sciences and Engineering Research Council of Canada (NSERC) for supporting this work.



%\bibliographystyle{IEEEtran}
%\begin{thebibliography}{1}
%\end{thebibliography}

\bibliographystyle{IEEEtran}
\setlength{\bibitemsep}{.2\baselineskip plus .05\baselineskip minus .05\baselineskip}

\bibliography{bibi}

\end{document}